\documentclass[10pt,twocolumn,letterpaper]{article}
\usepackage{geometry}
\usepackage{cvpr}
\usepackage{times}
\usepackage{epsfig}
\usepackage{graphicx}
\usepackage{amsmath}
\usepackage{amssymb}

\usepackage{amsfonts}
\usepackage{subfigure}
\usepackage{titling}
\pretitle{\begin{center}\Large \bf}
\posttitle{\par\end{center}}
\preauthor{\begin{center}\large\lineskip .5em \begin{tabular}[t]{c}}
\postauthor{\end{tabular}\par\vskip .5em \end{center}}
\newcommand{\modifier}[1]{{\color{black}#1}}
\newcommand{\erhao}{\fontsize{21pt}{\baselineskip}\selectfont}
\usepackage[breaklinks=true,bookmarks=false]{hyperref}

\cvprfinalcopy 

\def\cvprPaperID{4117} 

\ifcvprfinal\fi
\begin{document}

{\onecolumn

\noindent \textbf{\erhao{Deep Dual Relation Modeling for Egocentric Interaction Recognition}}

\vspace{2cm}

\noindent {\LARGE{Haoxin Li, Yijun Cai, Wei-Shi Zheng}}

%

\vspace{1cm}

\noindent For reference of this work, please cite:

\vspace{1cm}
\noindent Haoxin Li, Yijun Cai and Wei-Shi Zheng.
``Deep Dual Relation Modeling for Egocentric Interaction Recognition.''
In \emph{Proceedings of the IEEE Conference on Computer Vision and Pattern Recognition.} 2019.

\vspace{1cm}

\noindent Bib:\\
\noindent
@inproceedings\{li2019deepdual,\\
\ \ \   title=\{Deep Dual Relation Modeling for Egocentric Interaction Recognition\},\\
\ \ \  author=\{Li, Haoxin and Cai, Yijun and Zheng, Wei-Shi\},\\
\ \ \  booktitle=\{Proceedings of the IEEE Conference on Computer Vision and Pattern Recognition\},\\
\ \ \  year=\{2019\}\\
\}
}

%
\restoregeometry

\title{Deep Dual Relation Modeling for Egocentric Interaction Recognition}

\author{Haoxin Li\textsuperscript{1,3,4},
Yijun Cai\textsuperscript{1,4},
Wei-Shi Zheng\textsuperscript{2,3,4,}\thanks{Corresponding author}\\
\textsuperscript{1}{School of Electronics and Information Technology, Sun Yat-sen University, China}\\
\textsuperscript{2}{School of Data and Computer Science, Sun Yat-sen University, China}\\
\textsuperscript{3}{Peng Cheng Laboratory, Shenzhen 518005, China}\\
\textsuperscript{4}{Key Laboratory of Machine Intelligence and Advanced Computing, Ministry of Education, China}\\
\tt\small lihaoxin05@gmail.com,
caiyj6@mail2.sysu.edu.cn,
wszheng@ieee.org}

\date{}


\maketitle
\thispagestyle{empty}

\begin{abstract}
Egocentric interaction recognition aims to recognize the camera wearer's interactions with the interactor who faces the camera wearer in egocentric videos. In such a human-human interaction analysis problem, it is crucial to explore the relations between the camera wearer and the interactor. However, most existing works directly model the interactions as a whole and lack modeling the relations between the two interacting persons. To exploit the strong relations for egocentric interaction recognition, we introduce a dual relation modeling framework which learns to model the relations between the camera wearer and the interactor based on the individual action representations of the two persons. Specifically, we develop a novel interactive LSTM module\modifier{, the key component of our framework,} to 
explicitly model the relations between the two interacting persons based on their individual action representations, which are collaboratively learned with an interactor attention module and a global-local motion module. Experimental results on three egocentric interaction datasets show the effectiveness of our method and advantage over state-of-the-arts.
\end{abstract}

\section{Introduction}
\begin{figure}[t]
\centering
\subfigure[Invisibility of the camera wearer]{
\includegraphics[width=1.5in]{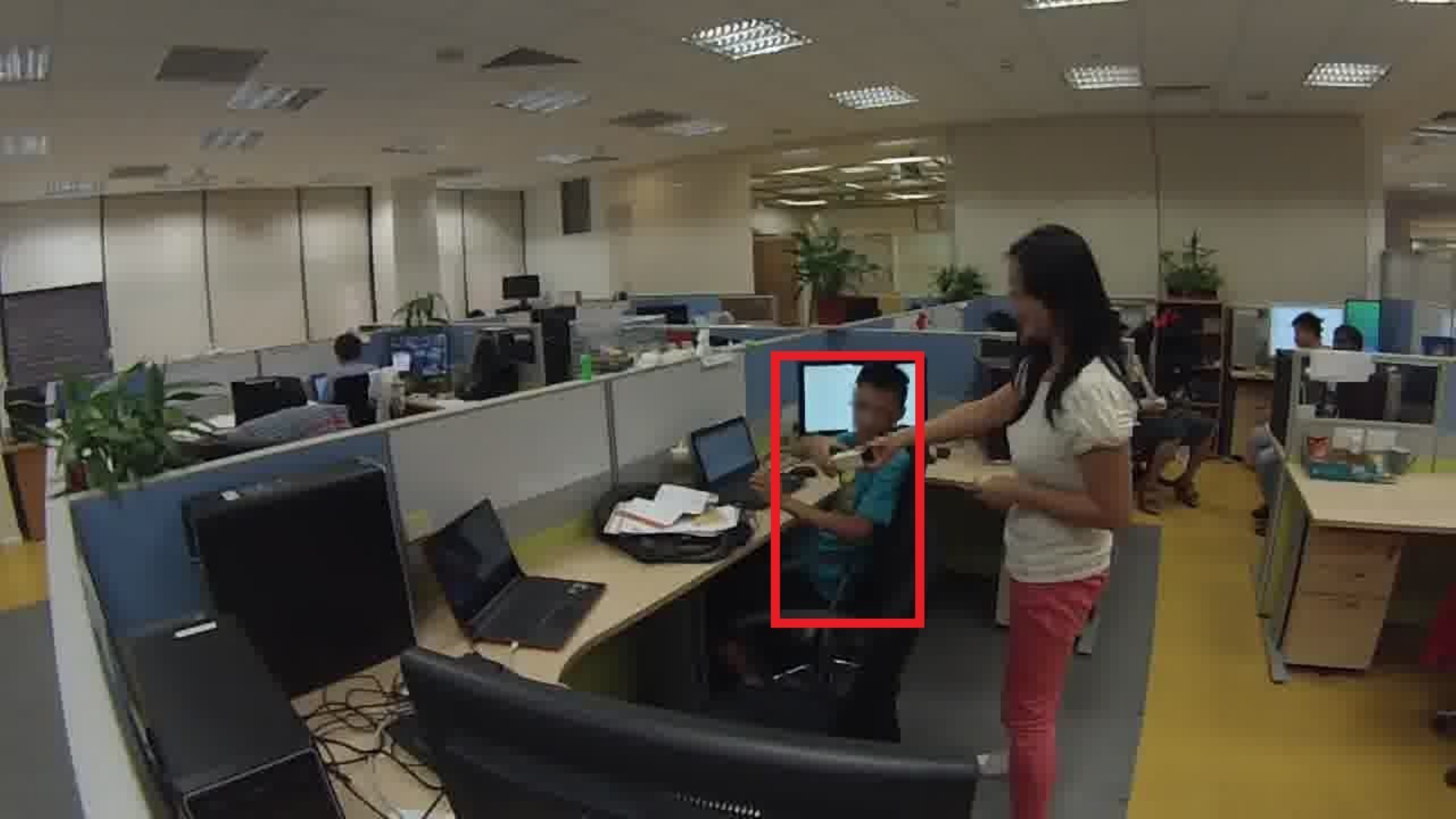}
\includegraphics[width=1.5in]{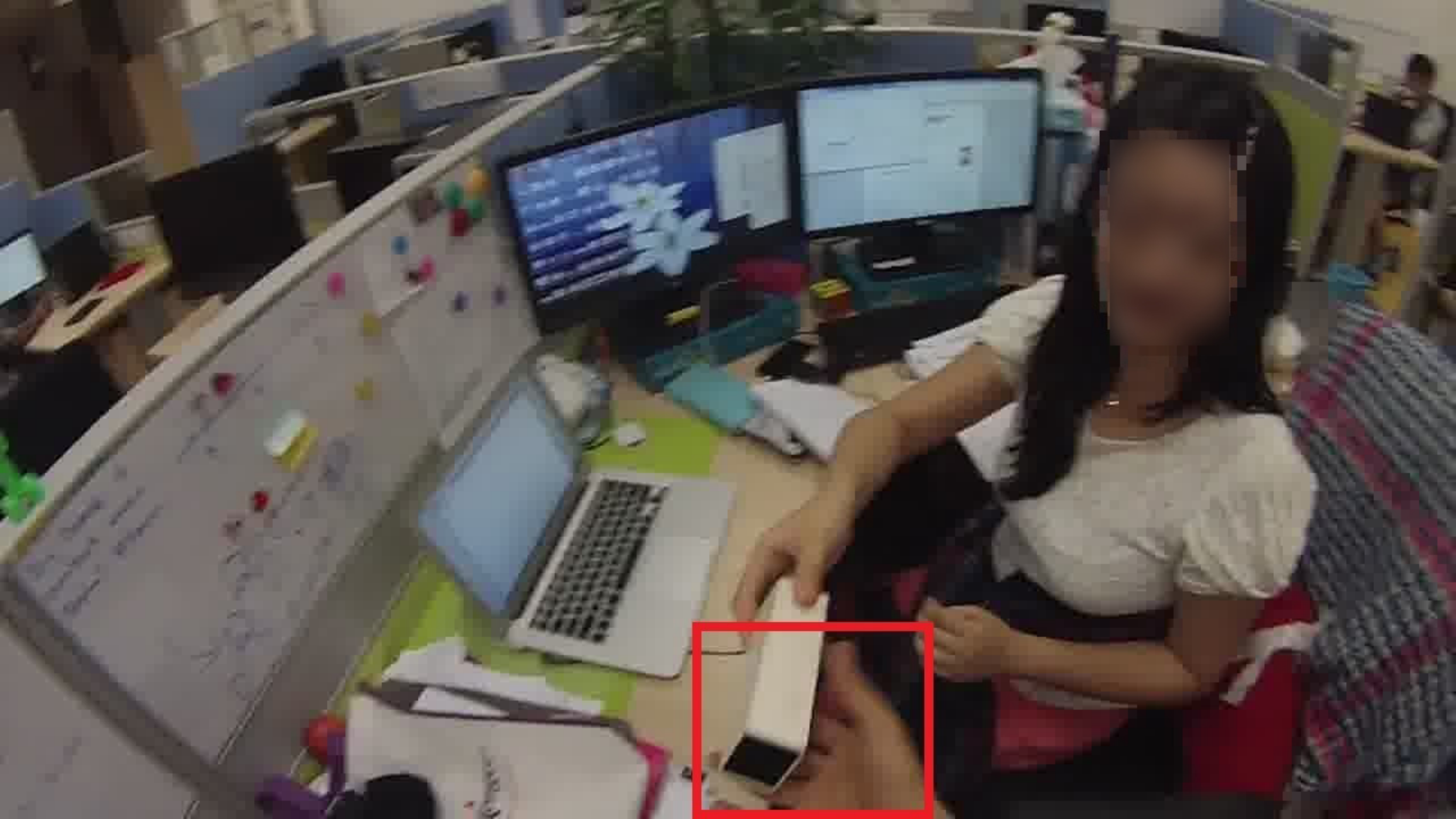}}
\subfigure[Ego-motion of the camera wearer]{
\includegraphics[width=1.5in]{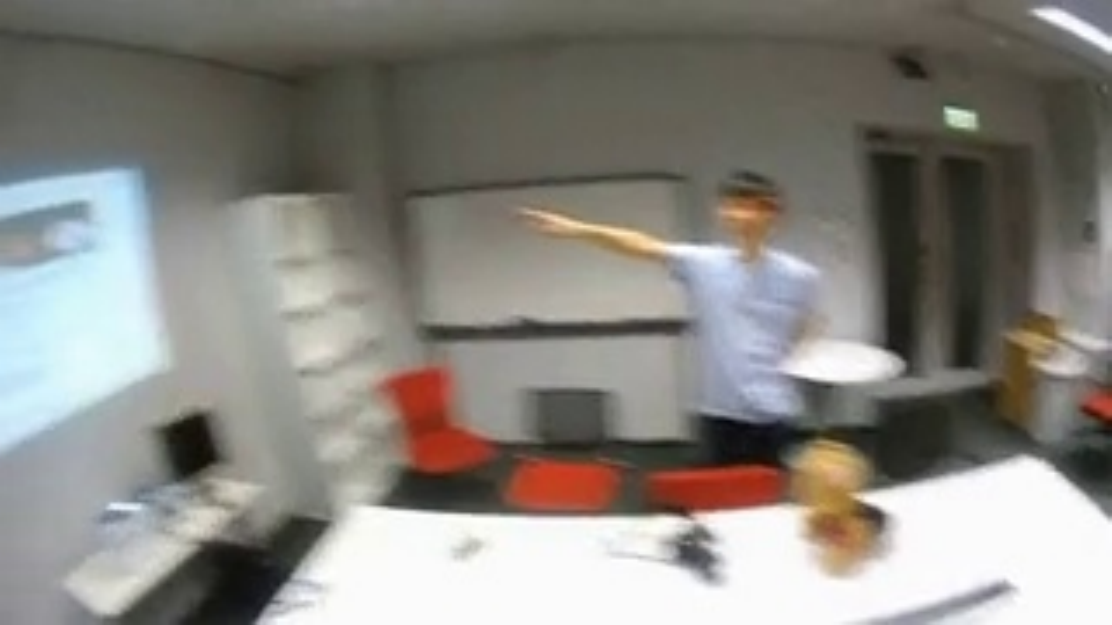}
\includegraphics[width=1.5in]{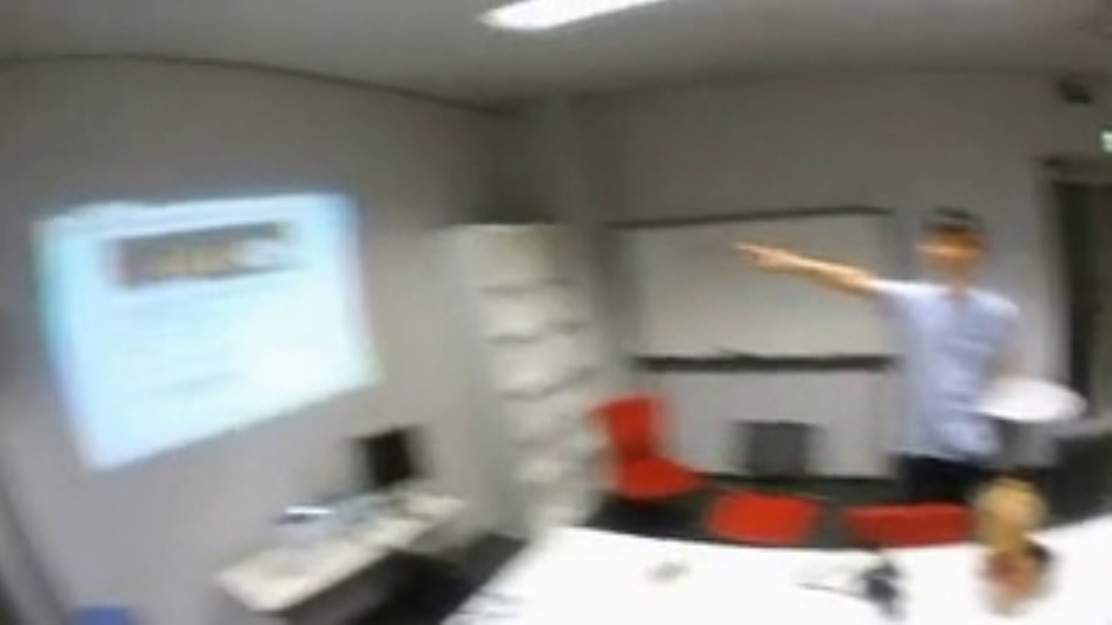}}
\caption{Illustration of camera-wearer's invisibility and ego-motion. (a) compares the person (in red boxes) receiving something in exocentric (left) and egocentric (right) videos from NUSFPID dataset \cite{narayan2014action}. (b) shows adjacent frames with obvious ego-motion in an egocentric video from UTokyo PEV dataset \cite{Yonetani_2016_CVPR}.}
\label{fig:challenge}
\end{figure}

Egocentric interaction recognition \cite{fathi2012social,narayan2014action,ryoo2013first,sudhakaran2017convolutional,Yonetani_2016_CVPR} attracts increasing attention with the popularity of wearable cameras and broad applications including human machine interaction \cite{NIPS2016_6113,8205953} and group events retrieval \cite{alletto2015understanding,alletto2014ego}. Different from exocentric (third-person) videos, in egocentric videos, the camera wearers are commonly invisible and the videos are usually recorded with dynamic ego-motion (see Figure \ref{fig:challenge}). The invisibility of the camera wearer \modifier{hampers action recognition learning of the camera wearer}, and the ego-motion \modifier{hinders direct motion description of the interactor, which make egocentric interaction recognition challenging.}

An egocentric interaction comprises the actions of the camera wearer and the interactor that influence each other with relations. So modeling the relations between the two interacting persons is important for interaction analysis. To model the relations between the the two interacting persons explicitly, we need to obtain individual action representations of the two persons \modifier{primarily}. Therefore, we formulate the egocentric interaction recognition problem into two interconnected subtasks, individual action representation learning and dual relation modeling.


In recent years, various works attempt to recognize interactions from egocentric videos. Existing methods integrated motion information using statistical properties of trajectories and optical flows \cite{narayan2014action,ryoo2013first,xia2015robot} or utilized face orientations descriptors \cite{fathi2012social} with SVM classifiers for recognition. Deep neural network was also adopted to aggregate short-term and long-term information for classification \cite{sudhakaran2017convolutional}. However, some of them \cite{ryoo2013first} aimed to recognize interactions from a static observer's view, which is impractical for most applications. Others \cite{fathi2012social,narayan2014action,sudhakaran2017convolutional} directly learned interaction as a whole through appearance and motion learning as done in common individual action analysis. They didn't learn individual action representations of the interacting persons, and thus failed to model the relations explicitly. The first-person and second-person features were introduced to represent the actions of the camera wearer and the interactor in \cite{Yonetani_2016_CVPR}. But they learned the individual action representations from multiple POV (point-of-view) videos and still lacked explicit relation modeling.

\begin{figure}[t]
\centering
\includegraphics[width=3.25in]{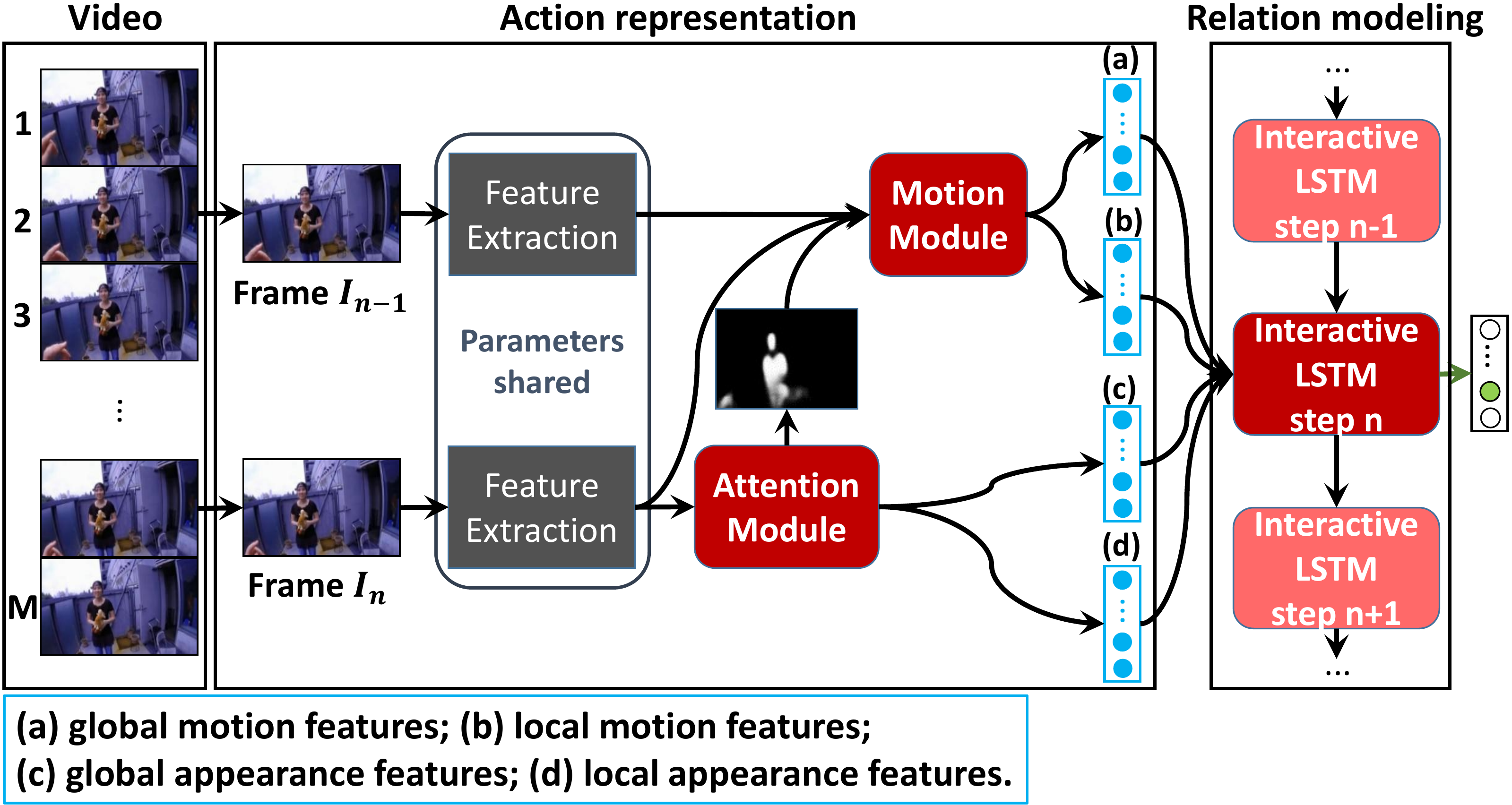}
\caption{Proposed framework. Frames $\boldsymbol{I}_{i}(i=1,...,N)$ are sampled from the video as input. The Feature Extraction Module extracts basic visual features of sampled frames. The Attention Module localizes the interactor and learns appearance features. The Motion Module estimates global and local motions for motion features learning. The Interaction Module models the relations for better interaction recognition based on the learned individual features (a), (b), (c) and (d) explained in the blue box.}
\label{fig:framework}
\end{figure}

\noindent
\textbf{Overview of the framework.} In this paper, we focus on the problem of recognizing human-human interactions from single POV egocentric videos. Considering the relations in egocentric interactions, we develop a dual relation modeling framework, which integrates the two interconnected subtasks, namely individual action representation learning and dual relation modeling, for recognition as shown in Figure \ref{fig:framework}. Specifically, for dual relation modeling, we develop an interaction module, termed interactive LSTM, to model the relations between the camera wearer and the interactor explicitly based on the learned individual action representations. For individual action representations learning, we introduce an attention module and a motion module to jointly learn action features of the two interacting persons. 
\modifier{We finally combine these modules into an end-to-end framework and train them with human segmentation loss, frame reconstruction loss and classification loss as supervision. Experimental results indicate the effectiveness of our method.}

\noindent
\textbf{Our contributions. }In summary, the main contribution of this paper is three-fold. (1) An interactive LSTM module is developed to model the relations between the camera wearer and the interactor from single POV egocentric videos. (2) An interactor attention module and a global-local motion module are designed to jointly learn individual action representations of the camera wearer and the interactor from single POV egocentric video. (3) By integrating individual action representations learning and dual relation modeling into \modifier{an end-to-end} framework, our method shows its effectiveness and outperforms existing state-of-the-arts on three egocentric interaction datasets.

\section{Related Work}
\noindent
\textbf{Egocentric action recognition} aims to recognize camera wearer's actions from first-person videos. Since the ego-motion is a dominant characteristic of egocentric videos, most methods used dense flow or trajectory based statistical features \cite{abebe2016robust,4587735,kitani2011fast,mi2018recognizing,7952632} to recognize the actions of the camera wearer. In some object-manipulated actions, some works extracted hands and objects descriptors for recognition \cite{6126269,li2015delving,6248010,7780579}, and others further explored gaze information according to hand positions and motions \cite{fathi2012learning,6751511}. Recently, deep neural networks have also been applied to egocentric action recognition. Frame-based feature series analysis showed their promising results \cite{kahani2017correlation,7298691,zaki2017modeling}. CNN networks with multiple information streams were also trained on recognition task \cite{ma2016going,singh2016first}. However, these methods target on individual actions which are a bit different from human-human interactions.

\noindent
\textbf{Egocentric interaction recognition} specifically focuses on first-person human-human interactions. Ryoo \etal recognized what the persons in the videos are doing to the static observer \cite{ryoo2015robot,ryoo2013first}, but it is unrealistic in most daily life scenarios. Some works used face orientations, individual locations descriptors and hand features to recognize interactions \cite{bambach2015lending,fathi2012social}. Others used motion information based on the magnitudes or clusters of trajectories and optical flows \cite{narayan2014action,xia2015robot}. A convLSTM was utilized to aggregate features of successive frames for recognition \cite{sudhakaran2017convolutional}. These methods commonly learned interaction descriptors by direct appearance or motion learning, but didn't considered explicit relation modeling with individual action representations of the camera wearer and the interactor. Yonetani \etal learned individual action features of the two persons but also lacked explicit relations modeling \cite{Yonetani_2016_CVPR}.

Different from the existing methods above, our framework jointly learns individual actions of the camera wearer and the interactor from single POV egocentric videos and further explicitly models the relations between them by an interactive LSTM.

\section{Individual Action Representation Learning}
To model the relations, we first need individual action representations of the camera wearer and the interactor. Here, we learn interactor masks to separate the interactor from background and learn appearance features with an attention module. \modifier{In the meanwhile, a motion module is integrated to learn motion cues, so that we can jointly learn the individual appearance and motion features of the two persons, which are the basis to model relations in Section \ref{sec:dual-relation}.}

For two consecutive sampled frames ${\boldsymbol{I}_{n-1}, \boldsymbol{I}_{n}}\in\mathbb{R}^{H\times W\times 3}$, we use a feature extraction module composed of ResNet-50 \cite{7780459} to extract basic features $\boldsymbol{f}(\boldsymbol{I}_{n-1}), \boldsymbol{f}(\boldsymbol{I}_{n})\in\mathbb{R}^{H_0\times W_0\times C}$, which encode the scene or human information with multidimensional representations for further modeling on top of them. In the following, we denote $\boldsymbol{f}(\boldsymbol{I}_{n-1})$ and $\boldsymbol{f}(\boldsymbol{I}_{n})$ as $\boldsymbol{f}^{n-1}$ and $\boldsymbol{f}^{n}$ respectively for convenience.

\subsection{Attention Appearance Features Learning}
\label{subsec:attention}
Egocentric videos record the actions of the camera wearer and the interactor simultaneously. To learn individual action features of the two persons, we wish to separate the interactor from background based on the feature $\boldsymbol{f}^{n}$.

\begin{figure}[t]
\centering
\includegraphics[width=3.0in]{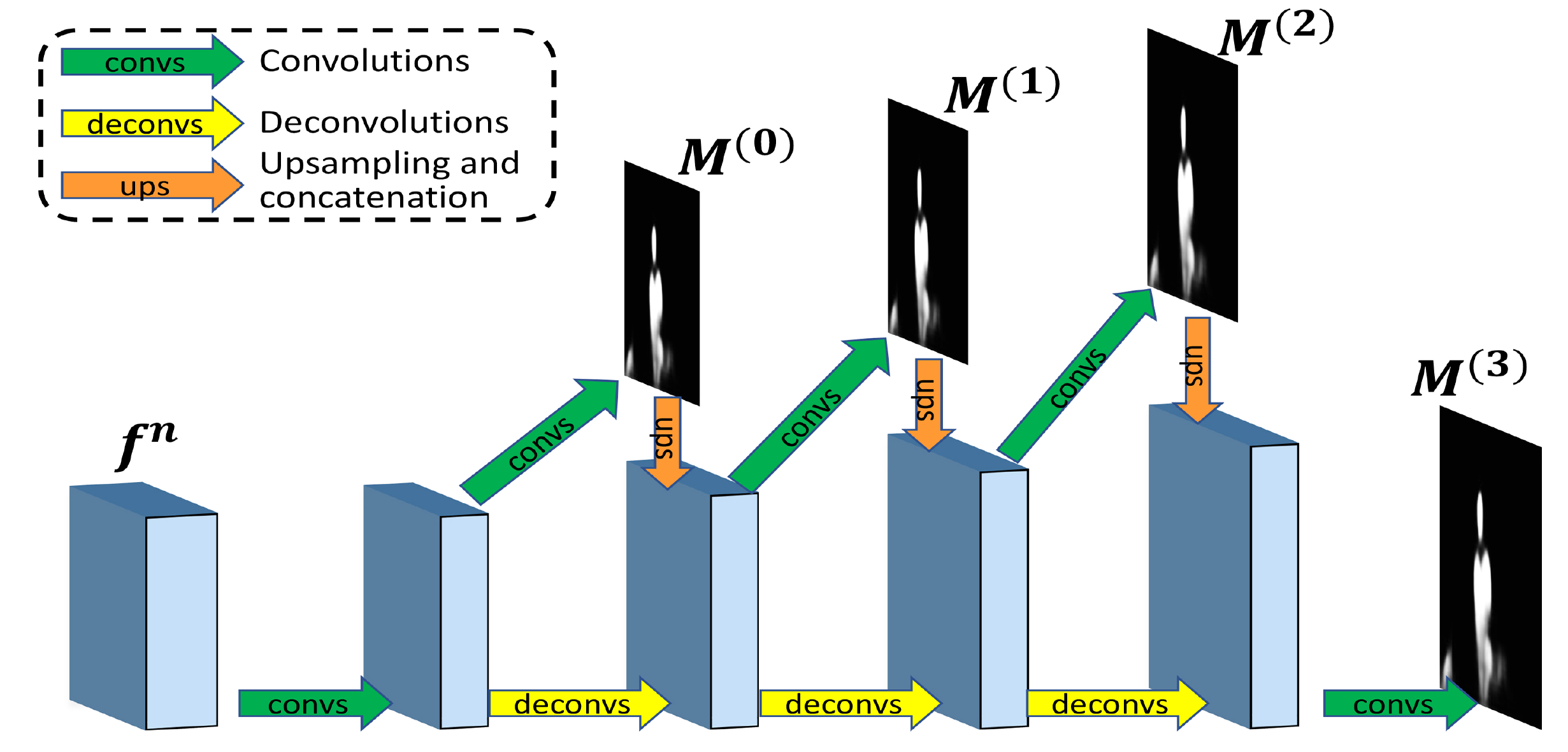}
\caption{Structure of attention module. The module takes feature $\boldsymbol{f}^n$ as input and generates attention weighted features and multi-scale masks.}
\label{fig:attention}
\end{figure}

Pose-guided or CAM-guided strategy \cite{du2017rpan,du2018recurrent,8300648} is used for person attention learning. Similarly, we introduce an attention module to localize the interactor with human segmentation guidance. We employ a deconvolution structure \cite{Noh_2015_ICCV} on top of the basic feature $\boldsymbol{f}^{n}$ to generate the masks of the interactor as shown in Figure \ref{fig:attention}. Mask $\boldsymbol{M}^{(0)}\in\mathbb{R}^{H_0\times W_0}$ serves to weight the corresponding feature maps for attention features learning. Multi-scale masks $\boldsymbol{M}^{(k)}\in\mathbb{R}^{H_k\times W_k}$($k=1,2,3$) are applied to localize the interactor at different scales for finer masks generation and explicit motion estimation later in Subsection \ref{subsec:motion}.

\noindent
\textbf{Mask of the Interactor.} To localize the interactor, we introduce a human segmentation loss to guide the learning of our attention module. Given a reference mask $\boldsymbol{M}^{RF}$, the human segmentation loss is a pixel-wise cross entropy loss:
\begin{equation}
\begin{split}
L_{seg}=&-\sum_{k=1}^{3}\sum_{i=1}^{H_k}\sum_{j=1}^{W_k}\frac{1}{H_k\times W_k}[\boldsymbol{M}^{RF}_{i,j}log\boldsymbol{M}^{(k)}_{i,j}+ \\ &(1-\boldsymbol{M}^{RF}_{i,j})log(1-\boldsymbol{M}^{(k)}_{i,j})],
\end{split}
\label{equ:segmentation-loss}
\end{equation}
where $k$ indexes the mask scales and the reference mask is resized to the corresponding shape for calculation. Here, the reference masks are obtained using JPPNet\cite{liang2018look}.

\noindent
\textbf{Attention Features.} An optimized attention module could localize the interactor, so the mask $\boldsymbol{M}^{(0)}$ has higher values at the positions corresponding to the interactor, which indicates concrete appearance information of the interactor. 
Then the {\bf{l}}ocal {\bf{a}}ppearance feature describing the action of the interactor from its appearance can be calculated with weighted pooling as follows:
\begin{equation}
\boldsymbol{f}_{l,a}^{n}=\frac{1}{|\boldsymbol{M}^{(0)}|}{\sum_{i=1}^{H_0}\sum_{j=1}^{W_0}\boldsymbol{M}^{(0)}_{i,j}\cdot \boldsymbol{f}^{n}_{i,j,1:C}},
\label{equ:local-appearance-feat}
\end{equation}
where $|\boldsymbol{M}^{(0)}|=\sum_{i=1}^{H_0}\sum_{j=1}^{W_0}\boldsymbol{M}^{(0)}_{i,j}$. Accordingly, the {\bf{g}}lobal {\bf{a}}ppearance feature, which describes the action of the camera wearer from what is observed, is calculated using global average pooling:
\begin{equation}
\boldsymbol{f}_{g,a}^{n}=\frac{1}{H_0\times W_0}{\sum_{i=1}^{H_0}\sum_{j=1}^{W_0}\boldsymbol{f}^{n}_{i,j,1:C}}.
\label{equ:global-appearance-feat}
\end{equation}


The attention module learns local appearance features \modifier{to provide a concrete description instead of a global description of the interactor}, and thus assists the relation modeling later. Meanwhile, the interactor masks localizing the interactor plays an important role in separating global and local motion, which is employed in Subsection \ref{subsec:motion}.

\subsection{Global-local Motion Features Learning}
\label{subsec:motion}
Motion features are vital for action analysis. To learn individual action representations of the two interacting persons, we wish to describe the ego-motion (global motion) of the camera wearer and the local motion of the interactor explicitly based on the basic features $\boldsymbol{f}^{n}$, $\boldsymbol{f}^{n-1}$ and the interactor masks $\boldsymbol{M}^{(k)}$($k=0,1,2,3$).

Differentiable warping scheme \cite{jaderberg2015spatial} is used for ego-motion estimation with a frame reconstruction loss \cite{vijayanarasimhan2017sfm,zhou2017unsupervised}. Inspired by them, we design a self-supervised motion module with the differentiable warping mechanism to jointly estimate the two types of motion from egocentric videos.

\noindent
\textbf{Global-local Motion Formulation by Reconstruction.} To separate the global and local motions in egocentric videos, we reuse the interactor mask $\boldsymbol{M}^{(3)}$ generated in Subsection \ref{subsec:attention} with the same scale as the input frames to formulate the transformation between two adjacent frames. With the learnable parameters $\boldsymbol{T}$ and $\boldsymbol{D}$ denoting transformation matrix and dense motion field, we can formulate the transformation from homogeneous coordinates $\boldsymbol{X}_n$ to $\boldsymbol{X}_{n-1}$ concisely as:
\begin{equation}
\boldsymbol{\hat{X}}_{n-1}=\boldsymbol{T}(\boldsymbol{X}_n+\boldsymbol{M}^{(3)}\odot\boldsymbol{D}),
\label{equ:trasformation}
\end{equation}
where $\odot$ is element-wise multiplication, $\boldsymbol{X}_n$ and $\boldsymbol{X}_{n-1}$ are homogeneous coordinates of frame $\boldsymbol{I}_n$ and $\boldsymbol{I}_{n-1}$.

In Equation (\ref{equ:trasformation}), $\boldsymbol{M}^{(3)}\odot\boldsymbol{D}$ is the local dense motion field of the interactor, and $\boldsymbol{T}$ describes the ego-motion of the camera wearer, so Equation (\ref{equ:trasformation}) jointly formulates global and local motion explicitly by point set reconstruction.

\noindent
\textbf{Self-supervision.} To learn the parameters in Equation (\ref{equ:trasformation}), we use view synthesis objective \cite{zhou2017unsupervised} as supervision:
\begin{equation}
L_{rec}=\sum_{x}{\vert \boldsymbol{I}_n(x)-\boldsymbol{\hat{I}}_n(x)\vert},
\label{equ:reconstruction-loss}
\end{equation}
where $x$ indexes over pixel coordinates $\boldsymbol{X}_n$. And $\boldsymbol{\hat{I}}_n$ is the reconstructed frame warped from frame $\boldsymbol{I}_{n-1}$ according to the transformed point set $\boldsymbol{\hat{X}}_{n-1}$, which is employed with the bilinear sampling mechanism \cite{jaderberg2015spatial} as
\begin{equation}
\boldsymbol{\hat{I}}_n(x)=\sum_{i\in\{t,b\},j\in\{l,r\}}w^{ij}\boldsymbol{I}_{n-1}(\hat{x}^{ij}),
\label{equ:bilinear-sampling}
\end{equation}
where $\hat{x}$ indexes over projected coordinates $\boldsymbol{\hat{X}}_{n-1}$, $\hat{x}^{ij}$ is the neighboring coordinate of $\hat{x}$, $w^{ij}$ is proportional to the spatial proximity between $\hat{x}^{ij}$ and $\hat{x}$, and subject to $\sum_{i,j}w^{ij}=1$. In addition, we regularize the local dense motions with a smoothness loss for robust learning \cite{vijayanarasimhan2017sfm}.

\begin{figure}[t]
\centering
\includegraphics[width=3.2in]{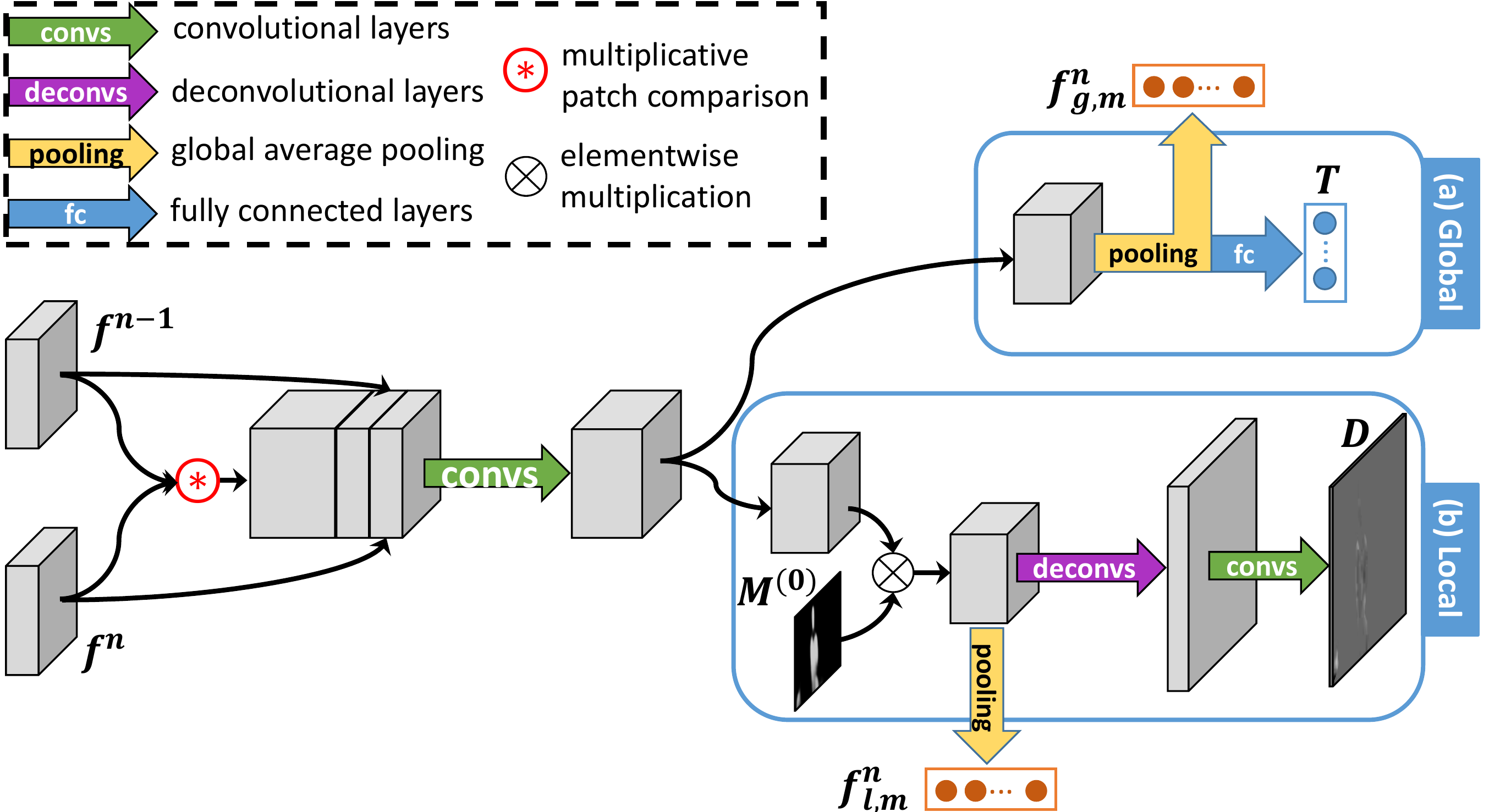}
\caption{Structure of motion module. The module takes basic features $\boldsymbol{f}^n$, $\boldsymbol{f}^{n-1}$ and mask $\boldsymbol{M}^{(0)}$ as inputs and estimates global and local motion parameters in two branches in which global motion feature $\boldsymbol{f}_{g,m}^{n}$ and local motion feature $\boldsymbol{f}_{l,m}^{n}$ are extracted. $\ast$ in red circle is a multiplicative patch comparison \cite{dosovitskiy2015flownet} to calculate the correlations between two feature maps, which captures the relative motions between them for dense flow estimation.}
\label{fig:motion}
\end{figure}

With the reconstruction loss in Equation (\ref{equ:reconstruction-loss}), we design a motion module illustrated in Figure \ref{fig:motion} with two branches learning the parameters of global ego-motion and local motion in Equation (\ref{equ:trasformation}), from which {\bf{g}}lobal {\bf{m}}otion feature $\boldsymbol{f}_{g,m}^{n}$ and {\bf{l}}ocal {\bf{m}}otion feature $\boldsymbol{f}_{l,m}^{n}$ are extracted from the embedding layers.

The motion module jointly estimates explicit motions of the camera wearer and the interactor by reusing the interactor masks, from which we learn concrete individual motion features of the two interacting persons hence aid relation modeling in Section \ref{sec:dual-relation}.

\subsection{Ego-feature and Exo-feature.}
\label{subsec:ego-exo-feat}
For each frame pair $\{\boldsymbol{I}_{n-1},\boldsymbol{I}_n\}$, we obtain global appearance feature $\boldsymbol{f}_{g,a}^{n}$ and local appearance feature $\boldsymbol{f}_{l,a}^{n}$ from the attention module, and also global motion feature $\boldsymbol{f}_{g,m}^{n}$ and local motion feature $\boldsymbol{f}_{l,m}^{n}$ from the motion module. The global features describe overall scene context and ego-motion of the camera wearer, which could represent the action of the camera wearer. While the local features, obtained with the interactor masks, describe the concrete appearance and motion of the interactor, which could represent the action of the interactor. Thus we obtain the individual action representations of the two interacting persons.

For further exploring the relations between the two persons, we define the \textbf{\emph{ego-feature }} $\boldsymbol{f}_{ego}^n=[\boldsymbol{f}_{g,a}^n, \boldsymbol{f}_{g,m}^n]$ describing the camera wearer, and \textbf{\emph{exo-feature}} $\boldsymbol{f}_{exo}^n=[\boldsymbol{f}_{l,a}^n, \boldsymbol{f}_{l,m}^n]$ describing the interactor. With them we could model the relations in Section \ref{sec:dual-relation}.

\section{Dual Relation Modeling by Interactive LSTM}
\label{sec:dual-relation}
Given the action representations, a classifier may be trained for recognition as done in most previous works. However, as discussed before, a distinguishing property of egocentric human-human interactions is the relations between the camera wearer and the interactor, which deserves further exploration for better interaction representations.

We notice that only the ego-feature or exo-feature may not exactly represent an interaction. For the example shown in Figure \ref{fig:effect-interaction}, two interactions consist of similar individual actions: the camera wearer turning his head and the interactor pointing somewhere. In this case, neither the features of any action can identify an interaction sufficiently. However, some relations would clearly tell the differences of the two interactions, such as the sequential orders and the motion directions of the individual actions. To utilize the relations for recognition, we develop an interaction module to model the relations between the two persons based on the ego-feature and exo-feature defined in Subsection \ref{subsec:ego-exo-feat}.


\subsection{Symmetrical Gating and Updating}
\modifier{To model the relations such as the synchronism or complementarity between the two interacting persons, we integrate their action features using LSTM structure.}

We define \textbf{\emph{ego-state}} $\boldsymbol{F}_{ego}^{n}$ and \textbf{\emph{exo-state}} $\boldsymbol{F}_{exo}^{n}$ to denote the latent states \modifier{till the $n$\emph{-th} step to encode the evolution of the two actions, which correspond to ego-feature and exo-feature introduced in Subsection \ref{subsec:ego-exo-feat}, respectively.} We wish to mutually incorporate the action context of each interacting person at each time step to explore the relations such as the synchronism and complementarity. Thus, we utilize exo-state to filter out the irrelevant parts, enhance the relevant parts and complement the absent parts of the ego-state. Meanwhile, the exo-state is also filtered, enhanced and complemented by the ego-state. This symmetrical gating and updating mechanism is realized with two symmetrical LSTM blocks where each block works as follows:
\begin{align}
\begin{split}
[\boldsymbol{i}^n;\boldsymbol{o}^n;\boldsymbol{g}^n;\boldsymbol{a}^n]=&\sigma(\boldsymbol{W}\boldsymbol{f}^n+\boldsymbol{U}\boldsymbol{F}^{n-1}+ \\ &\boldsymbol{J}^{n-1}+\boldsymbol{b}),
\end{split}\\
\boldsymbol{J}^{n}=&\phi(\boldsymbol{V}\boldsymbol{F}^{n}_*+\boldsymbol{v}),\\
\boldsymbol{c}^n=&\boldsymbol{i}^n \boldsymbol{a}^n+\boldsymbol{g}^n \boldsymbol{c}^{n-1},\\
\boldsymbol{F}^n=&\boldsymbol{o}^n tanh(\boldsymbol{c}^n).
\label{equ:symm-lstm}
\end{align}
Here, the input gate, output gate, forget gate and update candidate are denoted as $\boldsymbol{i}^n$, $\boldsymbol{o}^n$, $\boldsymbol{g}^n$ and $\boldsymbol{a}^n$ respectively. $\sigma$ is tanh activation function for update candidate and sigmoid activation function for other gates. $\boldsymbol{F}_*$ is the latent state from the dual block, $\phi$ is ReLU activation function, and $\boldsymbol{J}^{n}$ is the modulated dual state. $\{\boldsymbol{W},\boldsymbol{U},\boldsymbol{V},\boldsymbol{b},\boldsymbol{v}\}$ are parameters of each LSTM block.

It is noted that the current ego-state integrates the historical information of the ego-actions and also the exo-actions into itself, and vice versa for exo-state. The ego-state and exo-state describe the interaction from the view of the camera weearer and the interactor respectively. In this symmetrical gating and updating manner, the symmetrical LSTM blocks model the interactive relations instead of a raw combination of two actions.

\subsection{Explicit Relation Modelling}
Besides the symmetrical LSTM blocks introduced above for implicitly encoding the dual relations into the ego-state and exo-state, we further explicitly model the dual relation. To this end, we introduce relation-feature $\boldsymbol{r}^n$ to explicitly calculate the relations with a nonlinear additive operation on the ego-state and exo-state:
\begin{equation}
\boldsymbol{r}^{n}=tanh(\boldsymbol{F}_{ego}^{n}+\boldsymbol{F}_{exo}^{n}).
\end{equation}

With the relation-feature $\boldsymbol{r}^n$ at each time step, we further model the time variant relations with another LSTM branch to integrate the historical relations into the relation-states $\boldsymbol{R}^n$, which can be formulated as follows:
\begin{align}
[\boldsymbol{i}^n;\boldsymbol{o}^n;\boldsymbol{g}^n;\boldsymbol{a}^n]&=\sigma(\boldsymbol{W}\boldsymbol{r}^n+\boldsymbol{U}\boldsymbol{R}^{n-1}+\boldsymbol{b}),\\
\boldsymbol{c}^n&=\boldsymbol{i}^n \boldsymbol{a}^n+\boldsymbol{g}^n \boldsymbol{c}^{n-1},\\
\boldsymbol{R}^n&=\boldsymbol{o}^n tanh(\boldsymbol{c}^n).
\label{equ:relation-state}
\end{align}
In the equations above, the gates and parameters are similarly denoted as those in the symmetrical LSTM blocks. In Equation (\ref{equ:relation-state}), $\boldsymbol{R}^n$ integrates historical and current relations information to explicitly represent the relations of the two actions at n-th time step during the interaction.

\begin{figure}[t]
\centering
\includegraphics[width=3.2in]{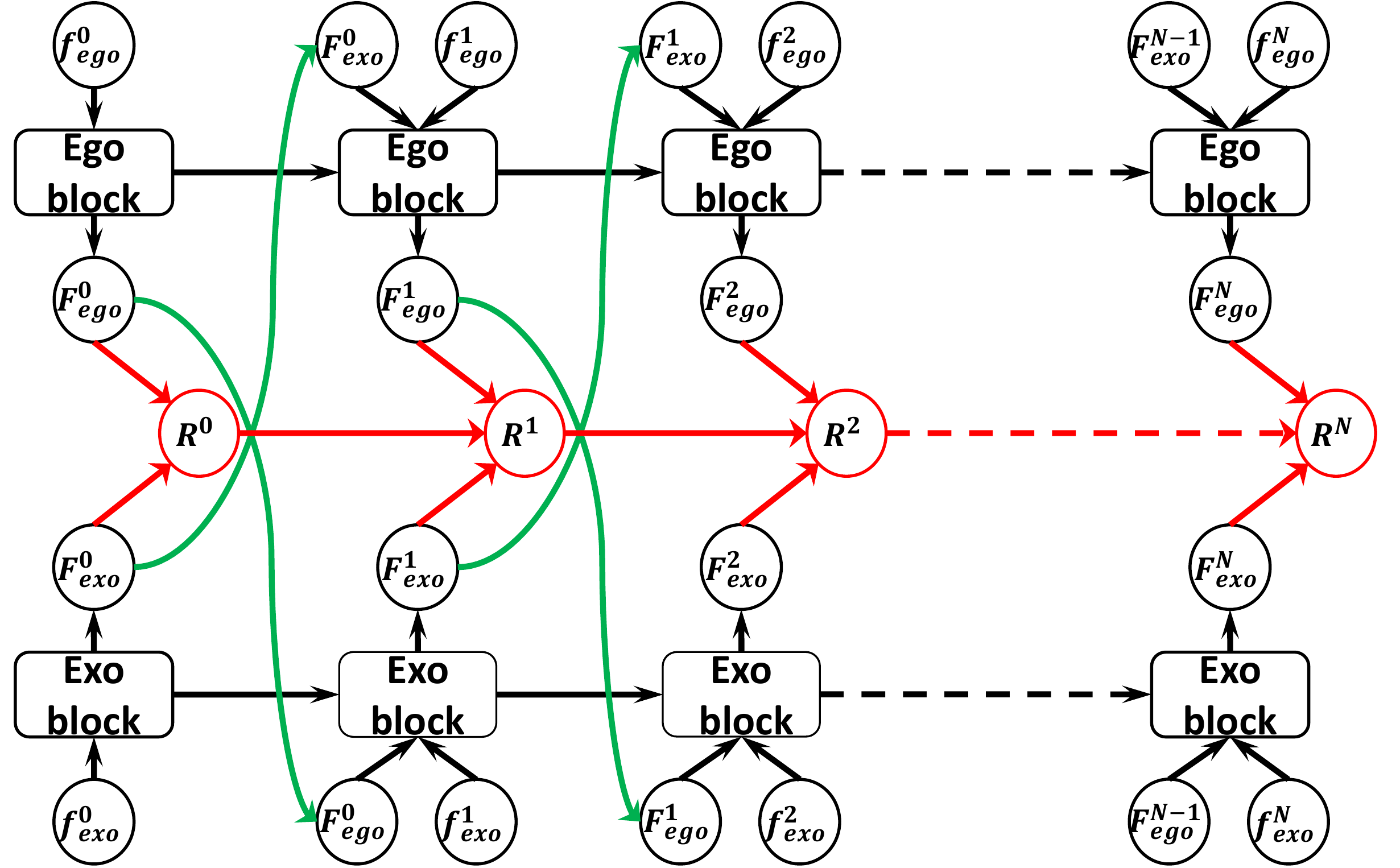}
\caption{Diagram of Interactive LSTM. The unrolled symmetrical LSTM blocks mutually gate and update each other as the green arrows depict. The unrolled relation LSTM branch is highlighted in red. All LSTM blocks contains $N$ time steps.}
\label{fig:interaction}
\end{figure}

Combining the two components above, \ie the symmetrical LSTM blocks and the relation LSTM branch, our interaction module is illustrated in Figure \ref{fig:interaction}, which we term interactive LSTM. It captures the evolution or synchronism of the two actions and further explicitly models the relations between the two actions, which provides a better representation of the interaction.

The posterior probability of an interaction category given the final relation-state $\boldsymbol{R}^N$ can be defined as
\begin{equation}
p(\boldsymbol{y}|\boldsymbol{R}^N)=\delta(\boldsymbol{W}\boldsymbol{R}^N+\boldsymbol{b}),
\label{equ:logits}
\end{equation}
where $\boldsymbol{W}$ and $\boldsymbol{b}$ are parameters of classifier and $\delta$ is softmax function. Then a cross entropy loss function is employed to supervise parameters optimization as follow:
\begin{equation}
L_{cls}=-\sum_{k=1}^{K}y_klog[p(y_k|\boldsymbol{R}^N)],
\label{equ:classification-loss}
\end{equation}
where $K$ is the number of class.

Combining the loss functions of each module above, we train our model end-to-end with the final objective:
\begin{equation}
L_{final}=L_{cls}+\alpha L_{seg}+\beta L_{rec}+\gamma L_{smooth},
\end{equation}
where $\alpha$, $\beta$, $\gamma$ are weights of segmentation loss, frame reconstruction loss and smooth regularization, respectively.

\section{Experiments}
\label{sec:experiment}

\begin{table*}[t]
\centering
\begin{tabular}{lcccc}
\hline
Methods&PEV&NUS(first h-h)&NUS(first)&JPL\\
\hline\hline
RMF\cite{abebe2016robust}&-&-&-&86.0 \\
Ryoo and Matthies\cite{ryoo2013first}&-&-&-&89.6 \\
Narayan \etal\cite{narayan2014action}&-&74.8&77.9&96.7 \\
Yonetani \etal\cite{Yonetani_2016_CVPR} (single POV)&60.4&-&-&75.0 \\
\hline
convLSTM\cite{sudhakaran2017convolutional} (raw frames)&-&-&69.4&70.6 \\
convLSTM\cite{sudhakaran2017convolutional} (difference of frames)&-&-&70.0&90.1 \\
LRCN\cite{7558228}&45.3&65.4&70.6&78.5 \\
TRN\cite{Zhou_2018_ECCV}&49.3&66.7&74.7&84.2 \\
Two-stream\cite{NIPS2014_5353}&58.5&78.6&80.6&93.4 \\
\hline
Our method&\bf{64.2}&\bf{80.2}&\bf{81.8}&\bf{98.4} \\
\hline
Yonetani \etal\cite{Yonetani_2016_CVPR} (multiple POV)&69.2&-&-&- \\
Our method (multiple POV)&69.7&-&-&- \\
\hline
\end{tabular}
\caption{State-of-the-art comparison (\%) with existing methods. \textit{NUS(first h-h)} denotes the first-person human-human interaction subset of NUS dataset and \textit{NUS(first)} denotes the first-person subset. It is notable that only PEV dataset provides multiple POV videos so that no multiple POV result of other datasets is reported.}
\label{tab:result}
\end{table*}

\subsection{Datasets}
We evaluate our method on three egocentric human-human interaction datasets.

\noindent
\textbf{UTokyo Paired Ego-Video (PEV) Dataset} contains 1226 paired egocentric videos recording dyadic human-human interactions \cite{Yonetani_2016_CVPR}. It consists of 8 interaction categories and was recorded by 6 subjects. We split the data into train-test subsets based on the subject pairs as done in \cite{Yonetani_2016_CVPR} and the mean accuracy of the three splits is reported.

\noindent
\textbf{NUS First-person Interaction Dataset} contains 152 first-person videos and 133 third-person videos of both human-human and human-object interactions \cite{narayan2014action}. We evaluate our method on first-person human-human interaction subset to verify the effectiveness of our method. To further test our method in human-object interaction cases, we also evaluate on the first-person subset. Random train-test split scheme is adopted and the mean accuracy is reported.

\noindent
\textbf{JPL First-Person Interaction Dataset} consists of 84 videos of humans interacting with a humanoid model with a camera mounted on its head \cite{ryoo2013first}. It consists of 7 different interactions. We validate our method's effectiveness in this static observer setting and report the mean accuracy over random train-test splits.

\subsection{Implementation Details}
\noindent
\textbf{Network Details.} In the motion module, we set 5 as the maximum displacement for the multiplicative patch comparisons. In the interaction module, we reduce the size of ego-feature and exo-feature to 256 and set 256 as the hidden size of LSTM blocks. 20 equidistant frames are sampled as input as done in \cite{sudhakaran2017convolutional}.

\noindent
\textbf{Data Augmentation.} We adopt several data augmentation techniques to ease overfitting due to the absence of large amount of training data. (1) Scale jittering \cite{wang2016temporal}. We fix the size of sampled frames as 160$\times$320 and randomly crop a region, then resize it to 128$\times$256 as input. (2) Each video is horizontally flipped randomly. (3) We adjust the hue and saturation in HSV color space of each video randomly. (4) At every sampling of a video, we randomly translate the frame index to obtain various samples of the same video.

\noindent
\textbf{Training setup.} The whole network is hard to converge if we train all the parameters together. Hence, we separate the training process into two stages. At the first stage, we initialize feature extraction module with ImageNet \cite{russakovsky2015imagenet} pretrained parameters and train attention module, motion module and interaction module successively while freezing other parameters. At the second stage, the three modules are finetuned together in an end-to-end manner. \modifier{We use Adam optimizer with initial learning rate $0.0001$ to train our model using TensorFlow on Tesla M40, and decrease the learning rate when the loss saturates. To deal with overfitting, we further employ large-ratio dropout, high weight regularization and early stop strategies during training.}

\subsection{Comparison to the State-of-the-art Methods}
We compare our method with state-of-the-arts and the results are shown in Table \ref{tab:result}. The first part lists the methods using hand-crafted features. The second part presents \modifier{some deep learning based action analysis methods (reimplemented by us except convLSTM)}. The third part reports the results of our method and the fourth part compares the performance using multiple POV videos on PEV dataset.

As shown, our method outperforms existing methods. Most previous methods directly learn interaction representations without relation modeling, while
\modifier{ours explicitly models the relations between the two interacting persons.} The results show that relation modeling is useful for interaction recognition.

Among the compared deep learning methods, we obtain clear improvement over convLSTM\cite{sudhakaran2017convolutional}, LRCN\cite{7558228} and TRN\cite{Zhou_2018_ECCV}, since they mainly capture the temporal changes of appearance features, but ours further explicitly captures motions and models the relation between the two interacting persons. Two-stream network \cite{NIPS2014_5353} with the same backbone CNN as ours integrates both appearance and motion features but obtains inferior performance to ours, perhaps due to the lack of relation modeling. 

On PEV dataset, Yonetani \etal\cite{Yonetani_2016_CVPR} achieves $69.2\%$ of accuracy with paired videos, certainly surpassing others using single POV video. We use our interactive LSTM to fuse the features from paired videos since there also exist some relations between the actions recorded by the paired videos. We achieve comparable result ($69.7\%$) which further proves the relation modeling ability of our interactive LSTM.

In terms of inference time, our framework takes around 0.15 seconds per video with 20 sampled frames, which is still towards real time. TRN\cite{Zhou_2018_ECCV} takes 0.04 seconds per video but it has clear lower recognition performance than ours. Although Two-stream\cite{NIPS2014_5353} obtains slightly inferior performance to ours, it takes 0.9 seconds per video since it spends much more time on extracting optical flows.

\subsection{Further Analysis}

\subsubsection{Study on Interaction Module.} 
Table \ref{tab:effect-interaction} compares recognition performance about interaction. It shows that our interactive LSTM clearly improves the performance, \modifier{since it models the relations and also drives feature learning of other modules.} 
On different datasets, the relation modeling obtains different performance gains. We obtain clearer improvements on PEV dataset since it contains more samples dependent on relations. While in NUS(first h-h) dataset, most samples have weaker relations between the two interacting persons.

\begin{table}[t]
\centering
\begin{tabular}{ccc}
\hline
Features&PEV&NUS(first h-h)\\
\hline\hline
Ego-features&55.2&67.9 \\
Exo-features&53.1&76.1 \\
Concat(no relation)&60.8&77.9 \\
Interaction with sym. blocks&62.7&78.1 \\
Interaction with rel. branch&63.0&79.0 \\
Interaction with both&64.2&80.2 \\
\hline
\end{tabular}
\caption{Recognition accuracy comparison (\%) about interacion. \textit{Concat(no relation)} means concatenation of ego-features and exo-features without any relation modeling. \textit{Interaction with sym. blocks} means only symmetrical LSTM blocks are used. \textit{Interaction with rel. branch} means only relation LSTM branch is used. \textit{Interaction with both} means both components are used.}
\label{tab:effect-interaction}
\end{table}

\begin{figure}[t]
\centering
\subfigure[Interaction category: none]{
\includegraphics[width=3.0in]{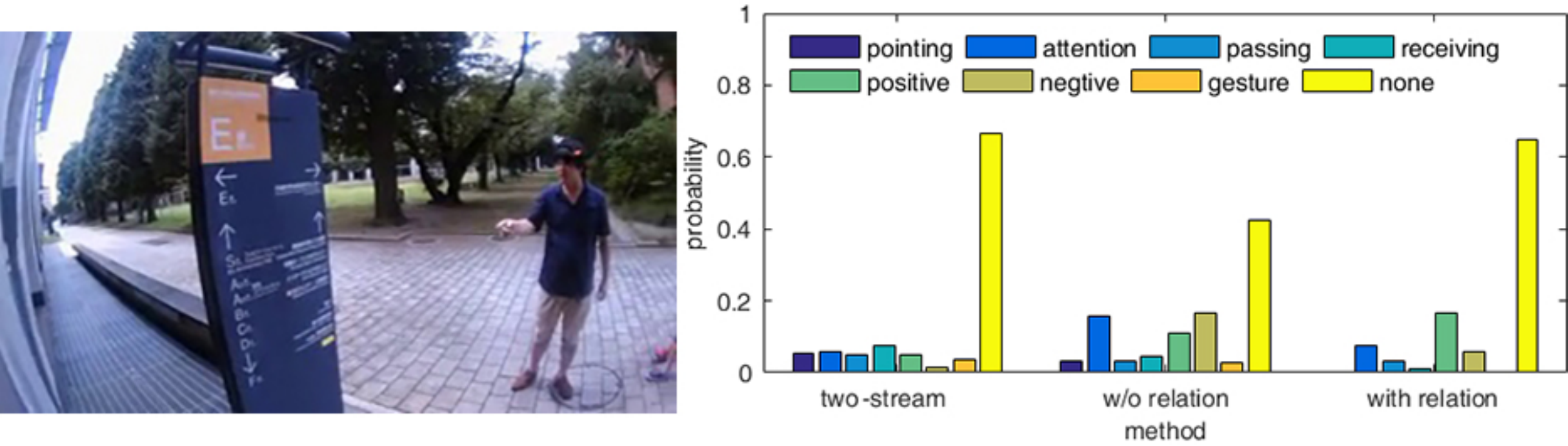}}

\subfigure[Interaction category: attention]{
\includegraphics[width=3.0in]{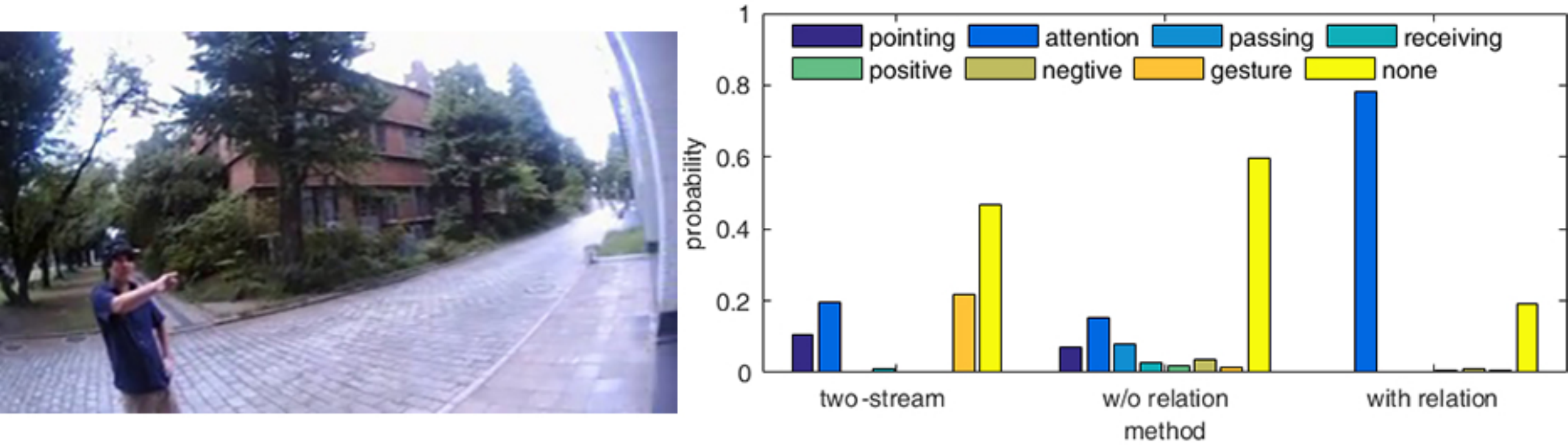}}

\caption{Comparison of recognition results of two interaction samples. \textit{w/o relation} means concatenation of two action features without any relation modelling is used for recognition. The bar graphs on the right present the probabilities of each category.}
\label{fig:effect-interaction}
\end{figure}

As discussed in Section \ref{sec:dual-relation}, the main differences between the two interaction samples shown in Figure \ref{fig:effect-interaction} might be the sequential orders and motion directions. We further compare the recognition results on them of different methods. It is observed that both two-stream\cite{NIPS2014_5353} and 
simple concatenation cannot sufficiently model the two interactions. While with explicit relation modeling, the two interactions are correctly distinguished, \modifier{which indicates that our interactive LSTM models the relations to distinguish confusing samples for better interaction recognition.}

\subsubsection{Study on Attention Module.}
We compare recognition accuracy of different appearance features in Table \ref{tab:effect-attention}. It is observed that local appearance features slightly improve the performance since it provides concrete descriptions of the interactor instead of general or overall features, which are more related to the interaction. Furthermore, relation modeling performs better than concatenation since it enhances the features through symmetrical gating or updating and relation modeling.

Figure \ref{fig:effect-attention} visualizes some learned masks. \modifier{(See more examples in supplementary material.)} As shown, the attention module learns to localize the interactor with the JPPNet reference masks as supervision. With additional classification loss, it could localize some objects around the interactor and strongly related to the interactions such as the hat in the example, which leads to around 2\% accuracy boost for local appearance features. This shows the advantage of using the designed attention module in our framework over using the JPPNet masks directly in this recognition task. In addition, with only the classification loss, our attention module fails to localize the interactor at all, indicating the necessity of reference masks for interactor localization.

The attention module is an indispensible part of our framework for individual action representation learning. It not only learns concrete appearance features, but also severs to separate the global and local motion for explicit motion features learning. Without attention module, our framework could only capture the global appearance and motion cues, and fails to model the relations between the camera wearer and the interactor, which leads to $9.0\%$ and $12.3\%$ accuracy degradation on PEV and NUS(first h-h) dataset, demonstrating the importance of attention module.

\begin{table}[t]
\centering
\begin{tabular}{ccc}
\hline
Features&PEV&NUS(first h-h)\\
\hline\hline
Two-stream\cite{NIPS2014_5353}(RGB)&40.7&63.8 \\
Global appearance&40.7&63.8 \\
Local appearance&43.2&65.1 \\
Concat(no relation)&44.2&66.8 \\
Interaction&45.9&68.2 \\
\hline
\end{tabular}
\caption{Recognition accuracy comparison (\%) using appearance features. \textit{Concat(no relation)} means simple concatenation of global and local appearance features. \textit{Interaction} means relations modeling is used with global and local appearance features.}
\label{tab:effect-attention}
\end{table}

\begin{figure}[t]
\centering
\subfigure[Frame]{
\includegraphics[width=1.5in]{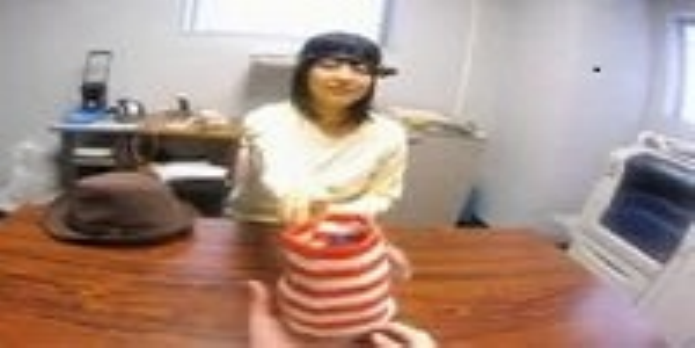}}
\subfigure[JPPNet mask]{
\includegraphics[width=1.5in]{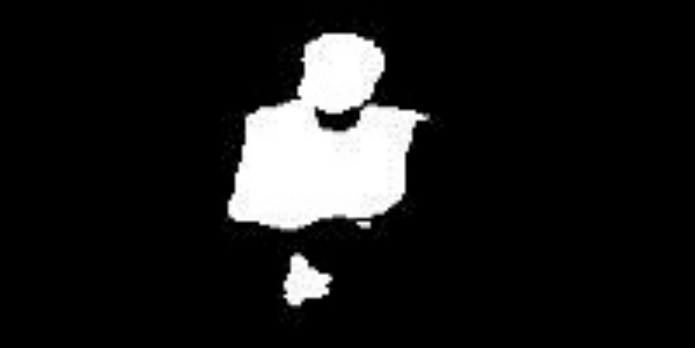}}

\subfigure[Mask]{
\includegraphics[width=1.5in]{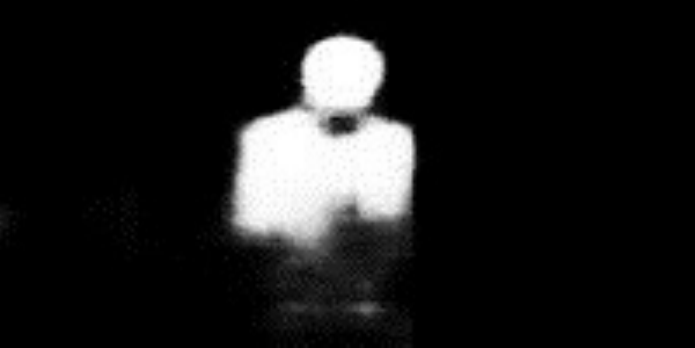}}
\subfigure[Mask]{
\includegraphics[width=1.5in]{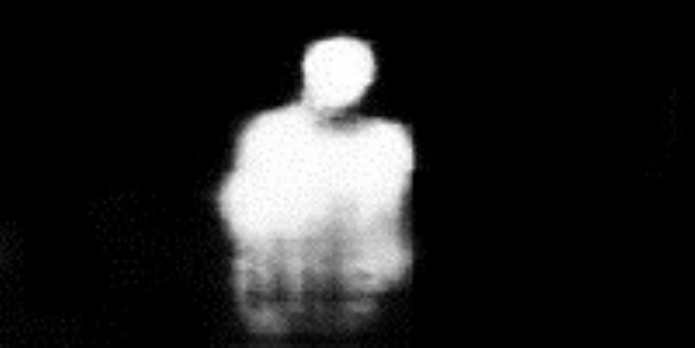}}
\caption{Example of learned mask with different supervision. (a) is original frame; (b) is the JPPNet mask; (c) is the learned mask trained with only human segmentation loss; (d) is the learned mask trained with human segmentation loss and classification loss.}
\label{fig:effect-attention}
\end{figure}

\subsubsection{Study on Motion Module.} 
We show accuracy comparisons of different motion features in Table \ref{tab:effect-motion}. It is seen that two-stream (flow) is a powerful method, but it is computational inefficient. Our method explicitly captures motions of the camera wearer and interactor and reaches comparable results with two-stream (flow), which indicates the effectiveness of our motion module. Furthermore, our method could achieve higher accuracy with relation modeling. On different datasets, global motion and local motion contribute differently to recognition, probably because global motion is important to distinguish interactions such as positive and negtive response on PEV dataset, but such interactions highly relevant to global motion are not included in NUS(first h-h) dataset.

\begin{figure}[t]
\centering
\subfigure[Frame $\boldsymbol{I}_{n-1}$]{
\includegraphics[width=1.5in]{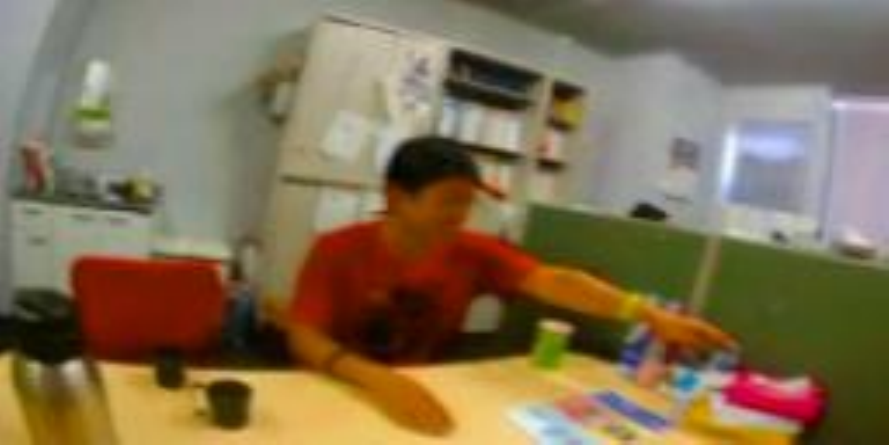}}
\subfigure[Frame $\boldsymbol{I}_n$]{
\includegraphics[width=1.5in]{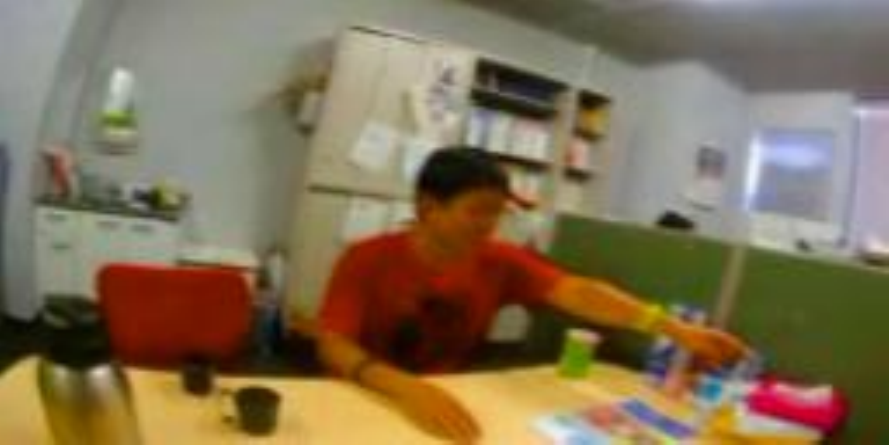}}

\subfigure[Local dense motion]{
\includegraphics[width=1.5in]{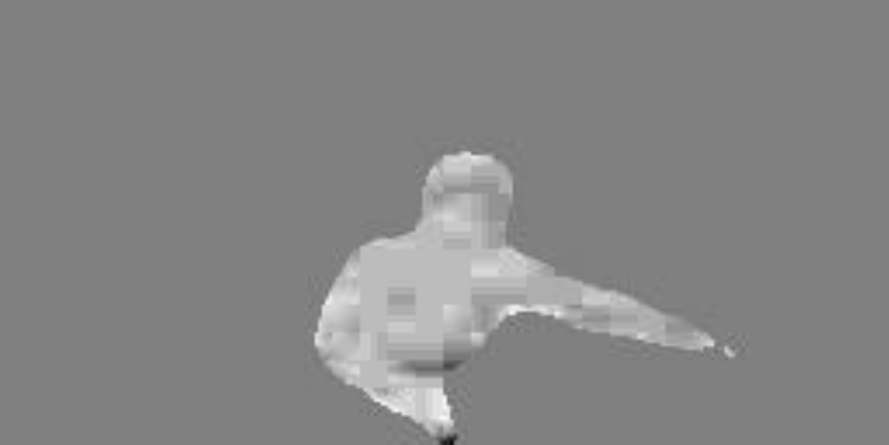}}
\subfigure[Global motion]{
\includegraphics[width=1.5in]{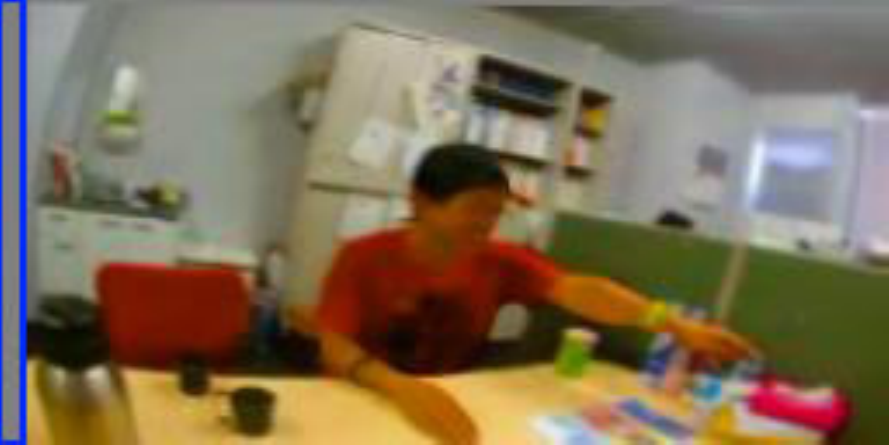}}
\caption{Illustration of global and local motion. (a) and (b) are two consecutive sampled frames. (c) Local dense motion shows the amplitudes of the horizontal motion vectors in the interactor mask, the amplitudes to the right are proportional to the brightness of the motion field. The motion vectors outside the interacor mask is discarded. (d) Global motion shows the slight head motion to the right, which reflects on the strip highlighted in blue.}
\label{fig:effect-motion}
\end{figure}

In Figure \ref{fig:effect-motion}, we show the reconstructed frame and local dense motion field. \modifier{(See more examples in supplementary material.)} From the reconstructed frame, it is seen that the the slight head motion to the right is captured, which leaves a strip on the left highlighted in blue. The local dense motion field shows the motion of the interactor reaching out the hand towards the right. This example shows that the motion module could learn the global and local motion jointly.

\begin{table}[t]
\centering
\begin{tabular}{ccc}
\hline
Features&PEV&NUS(first h-h)\\
\hline\hline
Two-stream\cite{NIPS2014_5353}(flow)&54.0&73.2 \\
Global motion&51.9&52.3 \\
Local motion&51.0&69.6 \\
Concat(no relation)&53.2&73.4 \\
Interaction&56.6&75.0\\
\hline
\end{tabular}
\caption{Recognition accuracy comparison (\%) using motion features. \textit{Concat(no relation)} means simple concatenation of global and local motion features. \textit{Interaction} means relations modeling is used with global and local motion features.}
\label{tab:effect-motion}
\end{table}

Our motion module explicitly estimates global and local motions of the camera wearer and the interactor individually, which is important for relation modeling. Without the motion module, our method fails to capture motion information and can only use appearance features, which leads to $18.3\%$ and $12.0\%$ accuracy drop on PEV and NUS(first h-h) dataset, showing the necessity of motion modeling.

\section{Conclusion}
In this paper, we propose to learn individual action representations and model the relations of the camera wearer and the interactor for egocentric interaction recognition. We construct a dual relation modeling framework by developing a novel interactive LSTM to explicitly model the relations. In addition, an attention module and a motion module are designed to jointly model the individual actions of the two persons for helping modeling the relations. Our dual relation modeling framework shows promising results in the experiments. \modifier{In the future, we would extend our method to handle more complex scenarios such as multi-person interactions, which are not considered in this paper.}

\section*{Acknowledgement}
This work was supported partially by the National Key Research and Development Program of China (2018YFB1004903), NSFC(61522115), and Guangdong Province Science and Technology Innovation Leading Talents (2016TX03X157).

{\small
\bibliographystyle{ieee_fullname}
\bibliography{egbib}
}

\title{Supplementary material for: \\
Deep Dual Relation Modeling for Egocentric Interaction Recognition}

\author{Haoxin Li\textsuperscript{1,3,4},
Yijun Cai\textsuperscript{1,4},
Wei-Shi Zheng\textsuperscript{2,3,4,*}\\
\textsuperscript{1}{School of Electronics and Information Technology, Sun Yat-sen University, China}\\
\textsuperscript{2}{School of Data and Computer Science, Sun Yat-sen University, China}\\
\textsuperscript{3}{Peng Cheng Laboratory, Shenzhen 518005, China}\\
\textsuperscript{4}{Key Laboratory of Machine Intelligence and Advanced Computing, Ministry of Education, China}\\
\tt\small lihaoxin05@gmail.com,
caiyj6@mail2.sysu.edu.cn,
wszheng@ieee.org}

\maketitle
\thispagestyle{empty}

\section*{FLOPs}
To evaluate the complexity of our model, we calculate the inference FLOPs (floating point operations) per video of variations of our model, the results are shown in Table \ref{tab:FLOPs-cmp}. It is observed that our relation modeling components would not increase FLOPs too much but still boost performance.
\vspace{-0.3cm}
\begin{table}[htb]
\centering
\caption{\footnotesize{Recognition accuracy (\%) and FLOPs of our model.}}
\label{tab:FLOPs-cmp}
\centering
\begin{tabular}{ccc}
\hline
Variations of model&PEV&inference FLOPs\\
\hline
Concat(no relation)&60.8&$8.4971\times10^{10}$ \\
Interaction with sym. blocks&62.7&$8.4974\times10^{10}$ \\
Interaction with rel. branch&63.0&$8.4982\times10^{10}$ \\
Interaction with both&64.2&$8.4984\times10^{10}$ \\
\hline
\end{tabular}
\end{table}

\section*{Visualization results}
To visualize the effect of our attention module and motion module, we illustrate some learned human masks and motions here. 

\begin{figure}[b]
\centering
\subfigure{\includegraphics[width=1.5in]{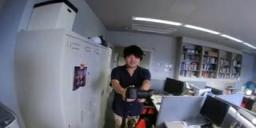}
\includegraphics[width=1.5in]{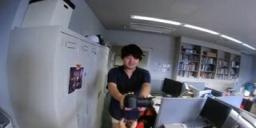}}

\vspace{-0.3cm} 
\subfigure{\includegraphics[width=1.5in]{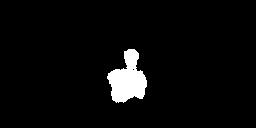}
\includegraphics[width=1.5in]{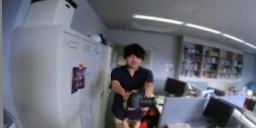}}

\vspace{-0.3cm} 
\subfigure{\includegraphics[width=1.5in]{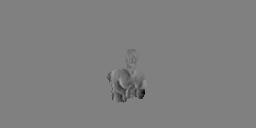}
\includegraphics[width=1.5in]{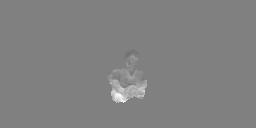}}
\caption{}
\label{fig:example1}
\end{figure}

\begin{figure}[b]
\centering
\subfigure{\includegraphics[width=1.5in]{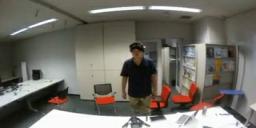}
\includegraphics[width=1.5in]{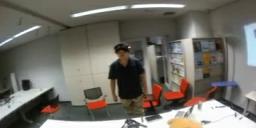}}

\vspace{-0.3cm} 
\subfigure{\includegraphics[width=1.5in]{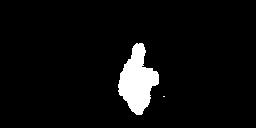}
\includegraphics[width=1.5in]{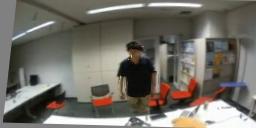}}

\vspace{-0.3cm} 
\subfigure{\includegraphics[width=1.5in]{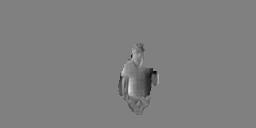}
\includegraphics[width=1.5in]{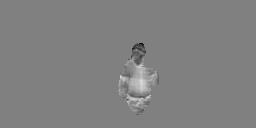}}
\caption{}
\label{fig:example2}
\end{figure}

\begin{figure}[h]
\centering
\subfigure{\includegraphics[width=1.5in]{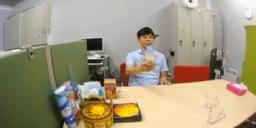}
\includegraphics[width=1.5in]{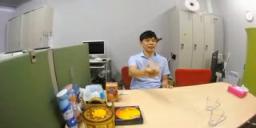}}

\vspace{-0.3cm} 
\subfigure{\includegraphics[width=1.5in]{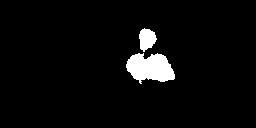}
\includegraphics[width=1.5in]{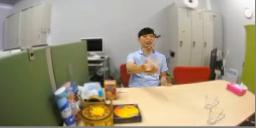}}

\vspace{-0.3cm} 
\subfigure{\includegraphics[width=1.5in]{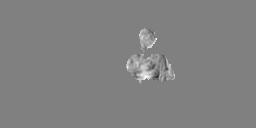}
\includegraphics[width=1.5in]{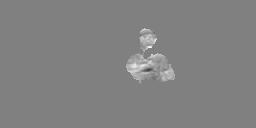}}
\caption{}
\label{fig:example3}
\end{figure}

\begin{figure}[h]
\centering
\subfigure{\includegraphics[width=1.5in]{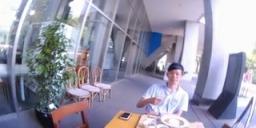}
\includegraphics[width=1.5in]{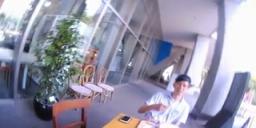}}

\vspace{-0.3cm} 
\subfigure{\includegraphics[width=1.5in]{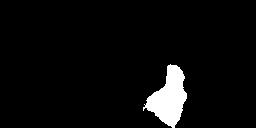}
\includegraphics[width=1.5in]{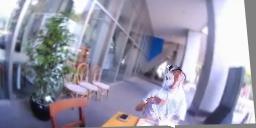}}

\vspace{-0.3cm} 
\subfigure{\includegraphics[width=1.5in]{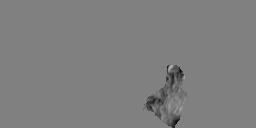}
\includegraphics[width=1.5in]{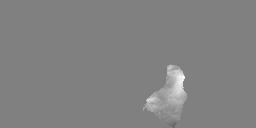}}
\caption{}
\label{fig:example4}
\end{figure}

\begin{figure}[h]
\centering
\subfigure{\includegraphics[width=1.5in]{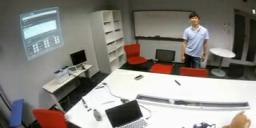}
\includegraphics[width=1.5in]{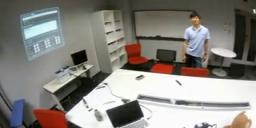}}

\vspace{-0.3cm} 
\subfigure{\includegraphics[width=1.5in]{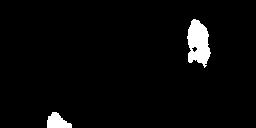}
\includegraphics[width=1.5in]{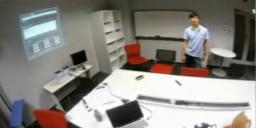}}

\vspace{-0.3cm} 
\subfigure{\includegraphics[width=1.5in]{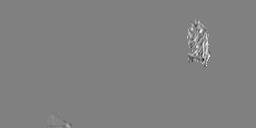}
\includegraphics[width=1.5in]{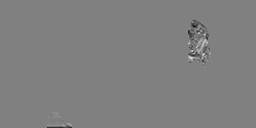}}
\caption{}
\label{fig:example5}
\end{figure}

\begin{figure}[h]
\centering
\subfigure{\includegraphics[width=1.5in]{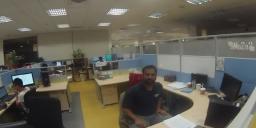}
\includegraphics[width=1.5in]{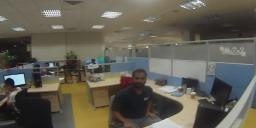}}

\vspace{-0.3cm} 
\subfigure{\includegraphics[width=1.5in]{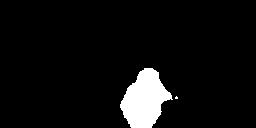}
\includegraphics[width=1.5in]{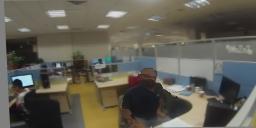}}

\vspace{-0.3cm} 
\subfigure{\includegraphics[width=1.5in]{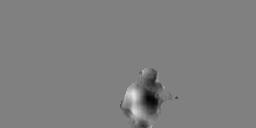}
\includegraphics[width=1.5in]{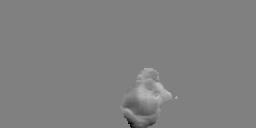}}
\caption{}
\label{fig:example6}
\end{figure}

\begin{figure}[h]
\centering
\subfigure{\includegraphics[width=1.5in]{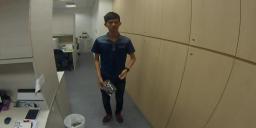}
\includegraphics[width=1.5in]{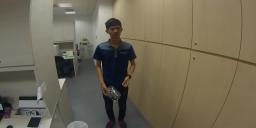}}

\vspace{-0.3cm} 
\subfigure{\includegraphics[width=1.5in]{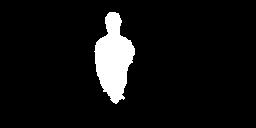}
\includegraphics[width=1.5in]{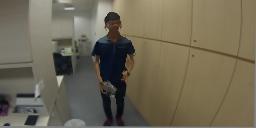}}

\vspace{-0.3cm} 
\subfigure{\includegraphics[width=1.5in]{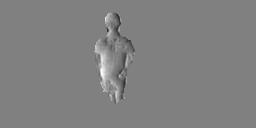}
\includegraphics[width=1.5in]{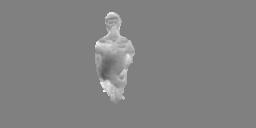}}
\caption{}
\label{fig:example7}
\end{figure}

\begin{figure}[h]
\centering
\subfigure{\includegraphics[width=1.5in]{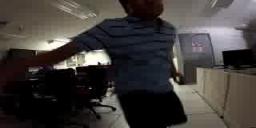}
\includegraphics[width=1.5in]{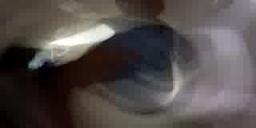}}

\vspace{-0.3cm} 
\subfigure{\includegraphics[width=1.5in]{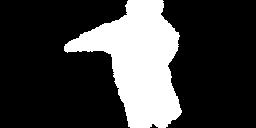}
\includegraphics[width=1.5in]{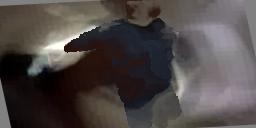}}

\vspace{-0.3cm} 
\subfigure{\includegraphics[width=1.5in]{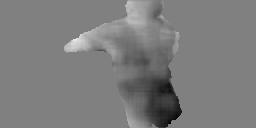}
\includegraphics[width=1.5in]{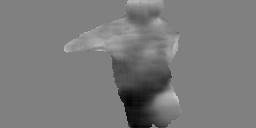}}
\caption{}
\label{fig:example8}
\end{figure}

Figure \ref{fig:example1} to Figure \ref{fig:example8} illustrate the learned human masks and motions of some samples. In each figure, the upper left and upper right are frame $\boldsymbol{I}_{n-1}$ and frame $\boldsymbol{I}_{n}$, respectively. The middle left is the learned human mask; the middle right is the reconstructed frame $\boldsymbol{\hat{I}}_{n-1}$ from frame $\boldsymbol{I}_{n}$ in which the global motion is reflected. The bottom left and the bottom right are the vertical and horizontal local motion field, respectively, where the amplitudes to the bottom and the right are proportional to the brightness of the motion fields. In addition, those motion vectors outside the learned human masks are discarded. \modifier{For example, in Figure \ref{fig:example1}, the reconstructed frame shows a slight head motion to the right, the vertical and horizontal local motion field together show the interactor's hands moving towards the upper right.}

It is observed that our attention module could learn to localize the interactor, and the motion module could capture the global and local motions in most cases,  except those with violent shakes as shown in Figure \ref{fig:example8}.

\end{document}